# Deep learning based single sample face recognition: a survey

**Fan Liu[1,2] · Delong Chen[1,2] · Fei Wang[1,2] · Zewen Li[1,2] · Feng Xu[1,2]**

**Abstract**
Face recognition has long been an active research area in the field of artificial intelligence, particularly since the rise of deep learning in recent years. In some practical situations, each identity has only a single sample available for training. Face recognition under this situation is referred to as single sample face recognition and poses significant challenges to the effective training of deep models. Therefore, in recent years, researchers have attempted to unleash more potential of deep learning and improve the model recognition performance in the single sample situation. While several comprehensive surveys have been conducted on traditional single sample face recognition approaches, emerging deep learning based methods are rarely involved in these reviews. Accordingly, we focus on the deep learning-based methods in this paper, classifying them into virtual sample methods and generic learning methods. In the former category, virtual images or virtual features are generated to benefit the training of the deep model. In the latter one, additional multi-sample generic sets are used. There are three types of generic learning methods: combining traditional methods and deep features, improving the loss function, and improving network structure, all of which are covered in our analysis. Moreover, we review face datasets that have been commonly used for evaluating single sample face recognition models and go on to compare the results of different types of models. Additionally, we discuss problems with existing single sample face recognition methods, including identity information preservation in virtual sample methods, domain adaption in generic learning methods. Furthermore, we regard developing unsupervised methods is a promising future direction, and point out that the semantic gap as an important issue that needs to be further considered.

Fan Liu and Delong Chen have contributed equally to this work.

✉ Fan Liu
fanliu@hhu.edu.cn

[1] College of Computer and Information, Hohai University, Nanjing, China

[2] Science and Technology on Underwater Vehicle Technology Laboratory, Harbin Engineering University, Harbin 150001, China

## 1 Introduction

As a convenient, natural, and highly accurate biometric recognition technology, face recognition has remained a hot research topic in the domains of artificial intelligence and computer vision since its inception. Significant progress has been made in this field over the past four decades. Recently, as powerful deep Convolutional Neural Networks (CNN) Li et al. (2021) have come to be applied to face recognition Wang and Deng (2021), their accuracy can potentially exceed even that of human beings. The advantage of deep learning-based face recognition is that deep models can learn to extract robust face features from large-scale training sets. However, in certain face recognition applications, such as identity card identification, passport recognition, judicial confirmation, and admission control, there is usually only one training sample available for each identity. In such situation, the problem is called single sample face recognition, or Single Sample Per Person (SSPP) face recognition Tan et al. (2006) and Kumar and Garg (2019). When traditional face recognition methods and deep methods encounter single sample problem, their performance will drop dramatically.

As shown in Fig. 1, single sample face recognition (blue) belongs to one-shot learning Fei-Fei et al. (2006) (green), which generally refers to a learning task with only a single labeled sample per class available. With the rise of deep learning, one-shot deep learning has gained more attention (Kadam and Vaidya 2020; O'Mahony et al. 2019). However, many deep one-shot methods cannot be applied directly to one-shot face recognition. This is because the inter-class variation in general one-shot tasks is large, while single sample face recognition is a fine-grained classification task with much smaller inter-class differences. Moreover, during testing, one shot learning methods widely follow the N-way one-shot setting Vinyals et al. (2016), where the value of N is usually set to 5 or 20. This type of testing strategy is very different from single sample face recognition, since single sample face recognition needs to compare hundreds of identities during testing.

In Fig. 2, we count the number of papers in the fields of single sample face recognition, one-shot learning, and deep learning published over the past 20 years. While, obviously, there have been numerous advances in deep learning and one-shot learning in recent years, but comparatively, not many novel methods have been proposed in the field of single sample face recognition. It is accordingly necessary to learn from the ideas and methods of deep learning and one-shot learning, then make specific adaptive improvements based on the unique characteristics of the face recognition task. This type of method is known as deep learning-based single sample face recognition, which is what this paper aims to review. Research in this field can not only improve the performance of the face recognition model in the single sample situation, but also inspire future research into one-shot learning and deep learning at the same time.

To the best of our knowledge, two surveys of single sample face recognition have already been published. In 2006, Tan et al. (2006) classified single sample face

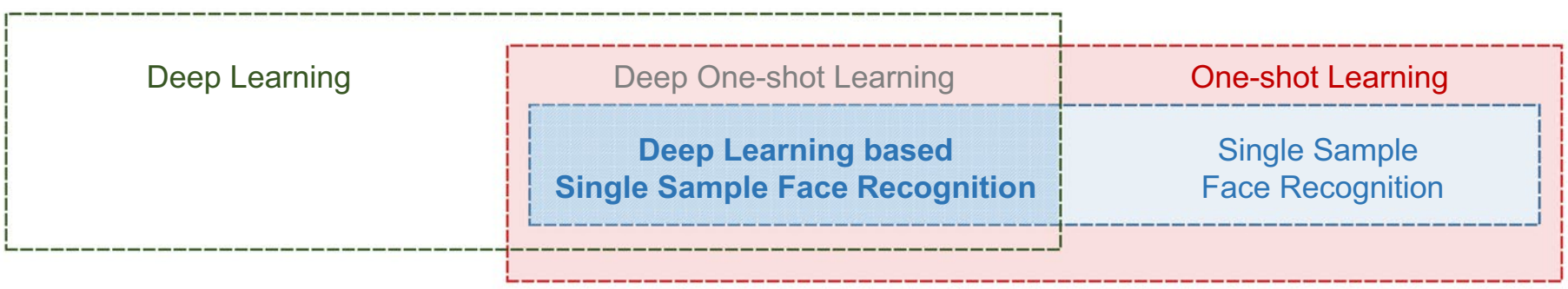

**Fig. 1** Relationships between researching fields

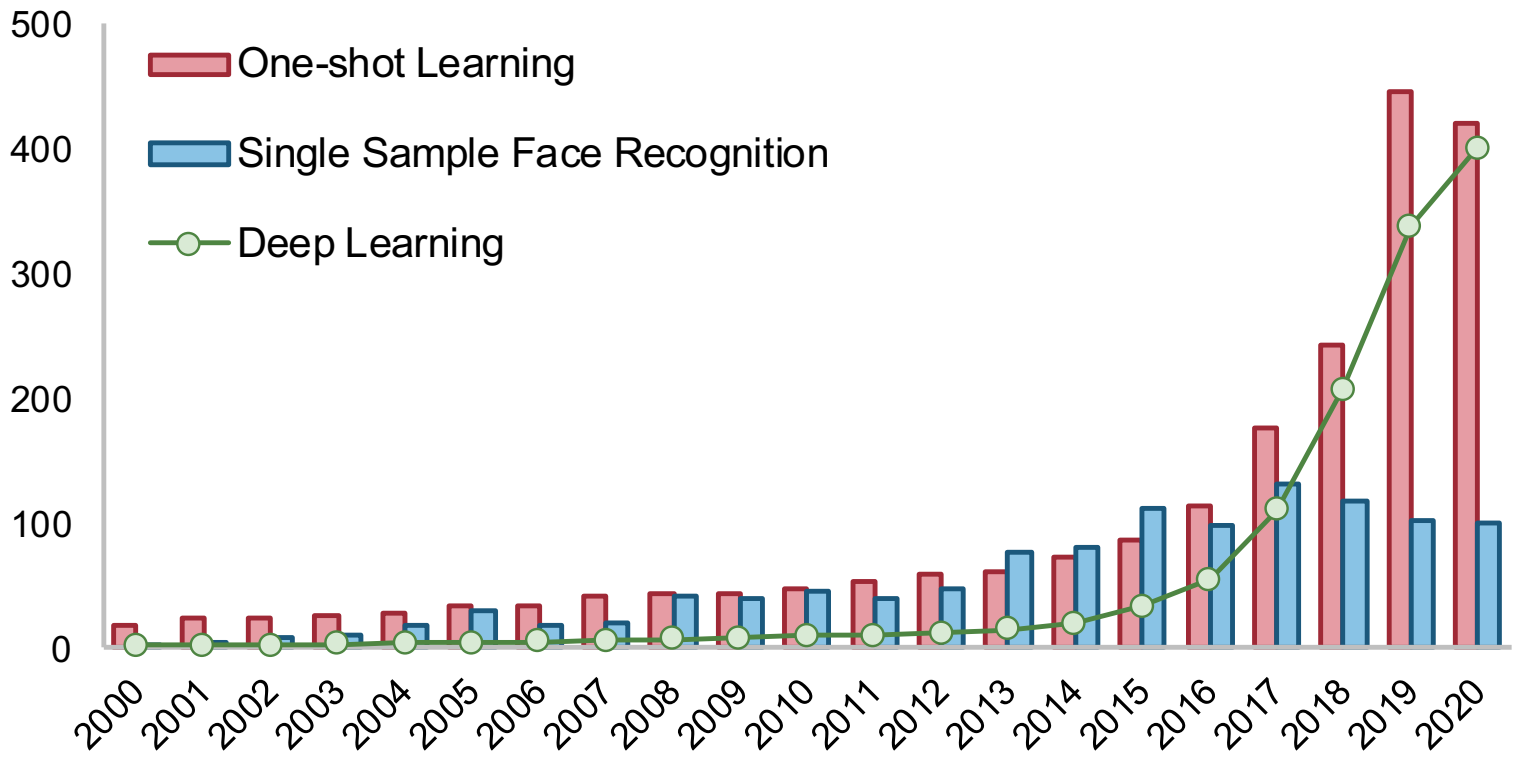

**Fig. 2** Number of papers published in research fields. The number of deep learning papers is too large so we divided them by 100 in this figure. Data is retrieved from Web of Science on Aug 20, 2021

recognition methods into three categories: global methods, local methods, and hybrid methods. These authors also compared and evaluated the performance of these methods. In 2019, Kumar and Garg (2019) summarized and reviewed the single sample face recognition methods that have emerged in the last decade, classifying them into feature-based methods, virtual sample methods, generic learning methods, hybrid methods, and other methods. These authors also compared the performance of the various methods and summarized the datasets for single sample face recognition. In addition to these, Li et al. (2018) surveyed virtual sample generation technology for small-sample-size face recognition in 2017. These authors categorized the virtual sample methods into face structure-based, perturbation and distribution function-based, and sample viewpoint-based methods. However, the surveys from Tan et al. (2006) and Li et al. (2018) focused on traditional face recognition methods only, and the survey from Kumar and Garg (2019) covered only two deep learning-based methods. Different from these survyes, in this paper, we primarily focus on deep learning-based single sample face recognition methods. The key contributions of this paper are as follows:

- We comprehensively reviewed existing deep learning based singe sample face recognition approaches, classifying them into virtual sample methods and generic learning methods. For methods in the former category, we classified them into virtual image generation and virtual feature generation; for those in the latter category, we classified them into those combining traditional methods and deep features, improving loss function, and improving network structure.
- We reviewed four popular testing datasets, including AR, Extend Yale B, LFW, and MS-Celeb-1M, then summarized and compared the performance of different types of models that have been tested on these datasets.
- We discussed common defects and deficiencies of existing methods. We analyzed the problems of identity information retention in virtual sample methods and domain adaptation in generic learning methods. Furthermore, we regarded large-scale unsupervised pre-training and the semantic gap as important issues that need to be considered in the future.

The rest of this paper is structured as follows. In Sect. 2, we conduct a detailed review and comparison of deep single sample face recognition approaches of each type. Datasets and performance comparisons are presented in Sect. 3, after which analysis and discussion are presented in Sect. 4. Finally, we conclude this paper in Sect. 5.

## 2 State of the art

Existing deep learning-based single sample face recognition methods can be divided into two types: virtual sample methods and generic learning methods. The development process of single sample face recognition is shown in Fig. 3. Applying conventional deep models directly to single sample face recognition often leads to model over-fitting due to the limited training samples. One direct approach to resolving this issue is generating virtual samples to enlarge the training set and convert the single sample face recognition task into a general multi-sample face recognition task; methods in this category are referred to as virtual sample methods. Another approach involves introducing an additional multi-sample generic set to improve the performance of the deep facial recognition model; methods in this category are referred to as generic learning methods.

### 2.1 Virtual sample methods

The key to the virtual sample methods is increasing the intra-class variation. Before the emergence of deep learning, many traditional methods were applied to the generation of virtual samples, including perturbation methods Martínez (2002), single image subspace method Liu et al. (2007), random sampling methods (Li et al. 2007; Vasilescu and Terzopoulos 2003), and decomposition and reconstruction methods (Li and Song 2011; Majumdar and Ward 2008). However, these methods generate virtual samples based on human-designed data augmentation rules, such that the resulting intra-class variation is limited. To better characterize the real variation of face images, researchers have proposed a variety of deep learning-based virtual sample methods in recent years. As shown in Figs 4 and 5, these methods can be divided into two categories: virtual image generation methods and virtual feature generation methods.

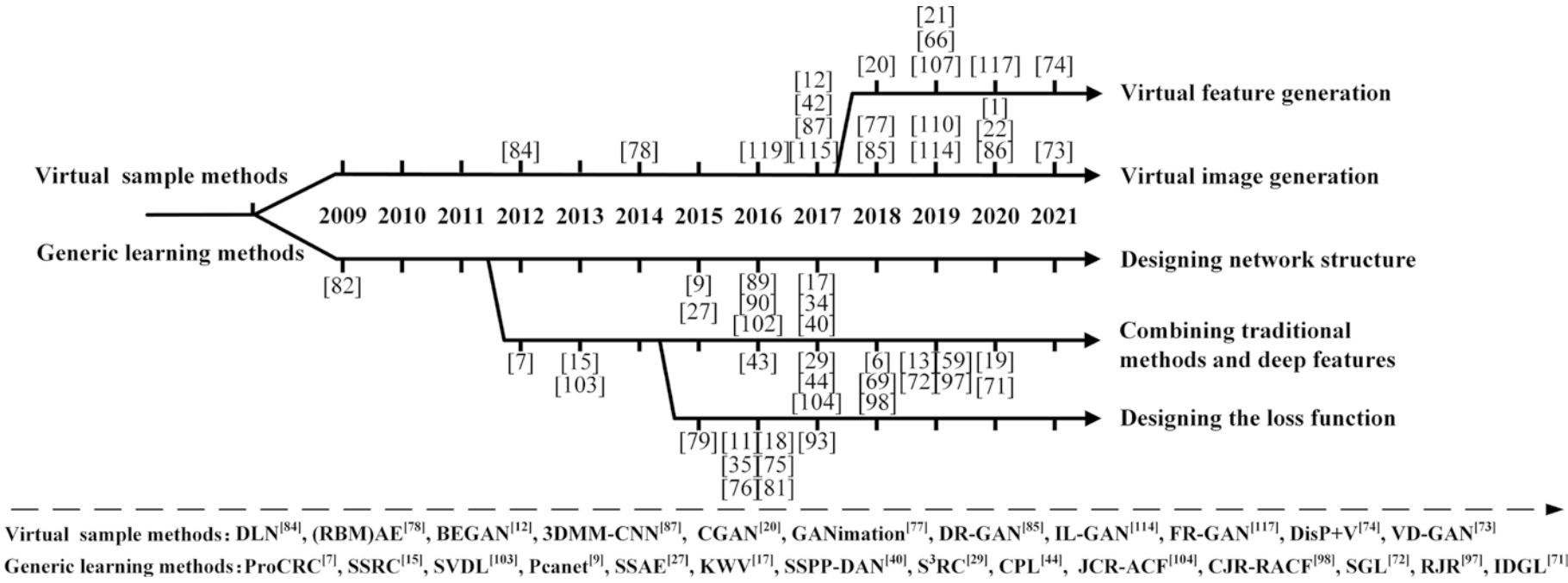

Fig. 3 The timeline of single sample face recognition methods

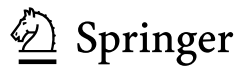

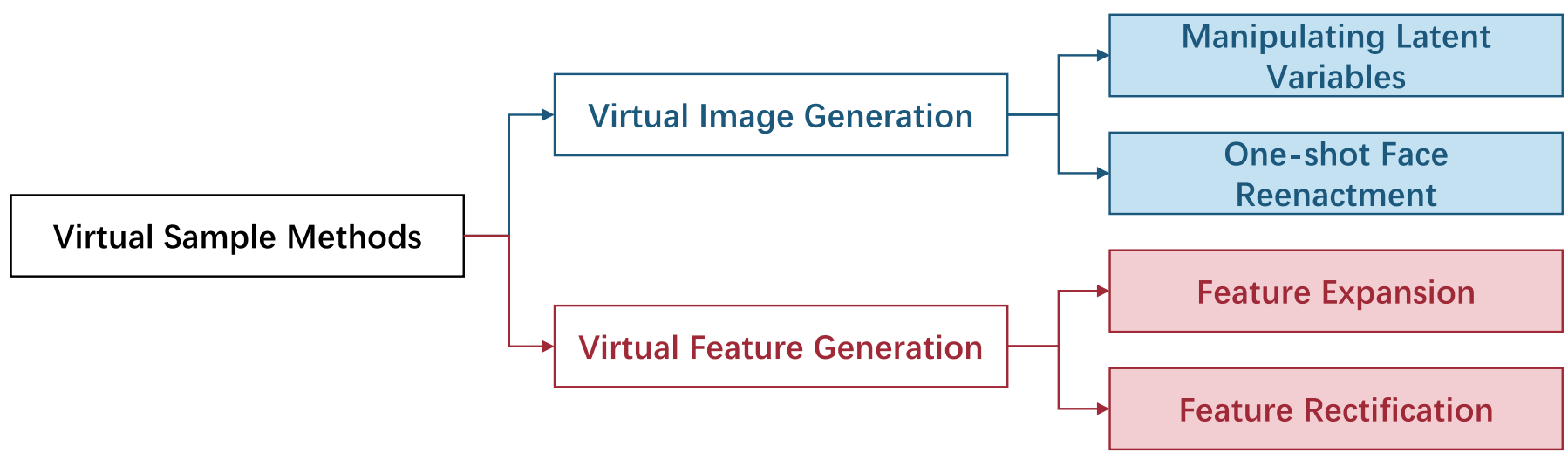

**Fig. 4** The different categories of virtual sample methods

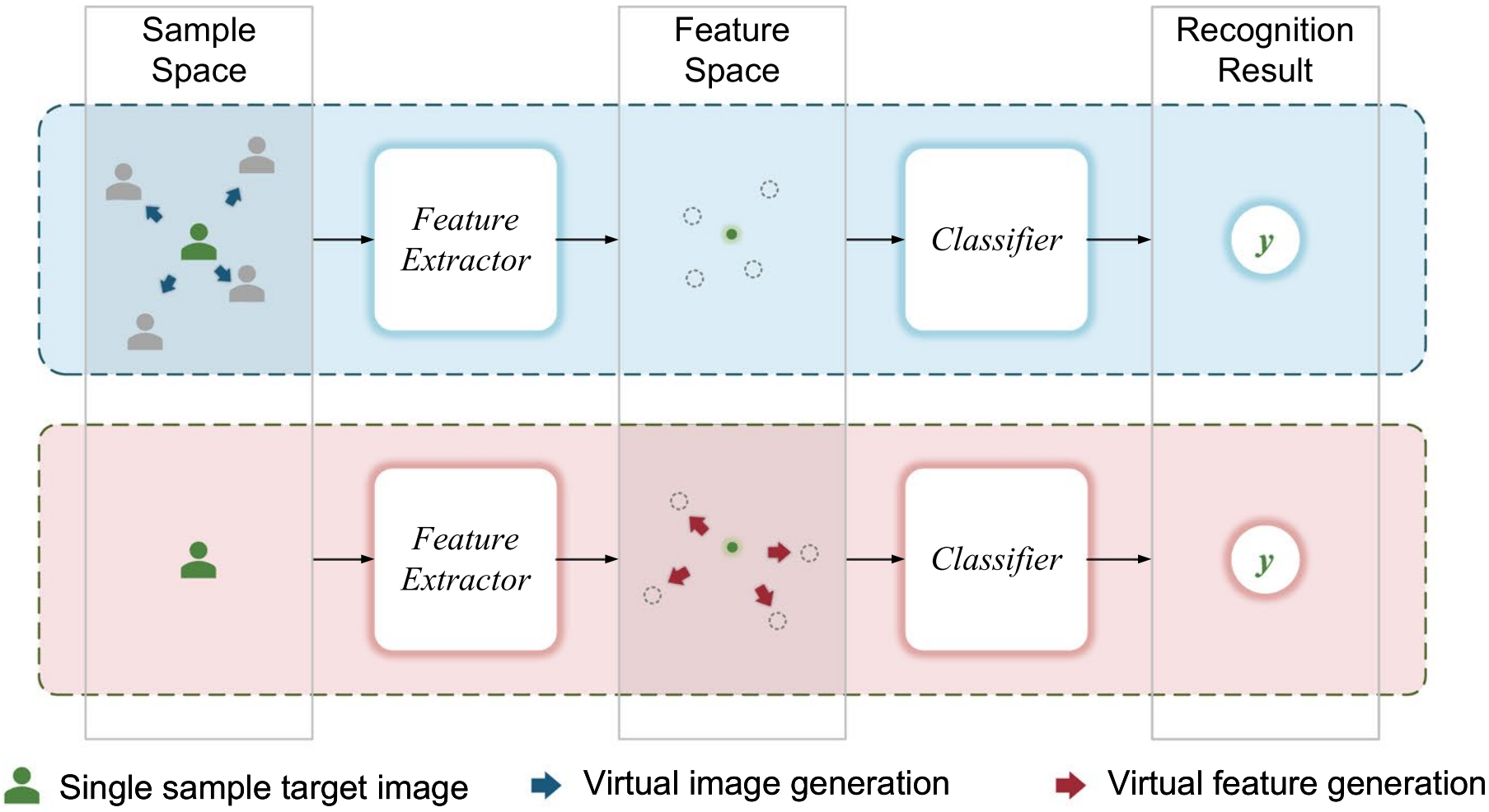

**Fig. 5** Diagrams of virtual image generation and virtual feature generation methods

### 2.1.1 Virtual image generation

Existing deep virtual image generation methods are based primarily on either AutoEncoder or Generative Adversarial Network (GAN). The AutoEncoder-based methods were developed earlier and have been tested more extensively. Deep learning was first introduced to virtual image generation in Abdolali and Seyyedsalehi (2010) and Abdolali and Seyyedsalehi (2012). The authors separated pose and identity components of latent variables in an unsupervised manner, then adjusted the pose component so as to enable the decoder to generate virtual samples of different poses. Reed et al. Reed et al. (2014) adopted Restricted Boltzmann Machine (RBM) and used a partially supervised approach to separate identity information and face variation. In these models, the identity information is represented by the extracted latent variables. As a different line of research, accurate 3D physical face models have also been applied to represent the identity information. For example, by using a 3D Morphable Model (3DMM) Blanz and Vetter (2003), Tuan et al. proposed an efficient 3DMM-CNN Tuan Tran et al. (2017) model that learns the parameters of the 3DMM using a deep convolutional neural network. Mokhayeri and Granger (2020) conducted a comparative face recognition performance evaluation of 3DMM and 3DMM-CNN models for virtual sample-based single sample face recognition. Their results show that while these two types of models achieve similar performance, 3DMM-CNN model requires far less processing time for virtual sample generation. Similarly, the Deep Lambertian Network (DLN) Tang et al. (2012) proposed by Hinton et al. assumes that the face is a Lambertian. Based on Deep Belief Net (DBN), they used the surface normal and the albedo to jointly represent the identity information and further used light angle vectors to represent intra-class variations. Following this idea, Tu et al. (2020) designed an AutoEncoder structure with one encoder and two decoders, corresponding to the shape and albedo reconstruction tasks respectively.

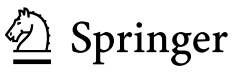

The above AutoEncoder-based models use an "encode—separate and adjust latent variables—decode" generation process. Some other methods use novel network structures and a different generation process. Zhu et al. (2014) introduced Multi View Perceptron (MVP) that can disentangle identity and view information, and generate virtual samples in different views. Zhang and Peng (2017) trained the AutoEncoder with an auxiliary dataset, then migrated the intra-class variation to the single sample. To ensure the generation quality and that the identity information is consistent, the intra-class variations are derived from neighbor samples of the single sample, which may limit the diversity of generated samples.

Furthermore, several other methods generated virtual samples with GANs. For example, Zakharov et al. (2019) used the extracted face feature points to represent the intra-class variation, then a meta-learning strategy is used to ensure that the model generates high-quality virtual samples in adversarial training. Tran et al. (2018) proposed a Disentangled Representation learning-Generative Adversarial Network (DR-GAN) for pose-controllable face generation. The discriminator of the DR-GAN simultaneously ensures the realism, identity preservation and pose correctness of generated face images. Choe et al. (2017) proposed a virtual sample generation method based on Boundary Equilibrium Generative Adversarial Networks (BEGAN). To obtain the virtual samples, this method inputs specifically adjusted latent variables into the trained BEGAN generator. However, the interpolation method of the mirrored faces' latent variables for pose transition may cause the identity information in the obtained virtual sample to change, especially for faces with asymmetry characteristics.

Another line of research is to integrate AutoEncoder and GAN. For example, Huang et al. (2017) and Duan and Zhang (2020) combined pixel-wise reconstruction loss, adversarial loss, and symmetry loss to train the face generator. The former two losses can be viewed as a direct combination of AutoEncoder and GAN, but the latter symmetry loss may leed to the identity information inconsistent problem similar to an aforementioned method Choe et al. (2017). Abdelmaksoud et al. (2020) sequentially leveraged GANs for superresolution and deblur, then applied an AutoEncoder structured model called Position map Regression Network (PRN) Feng et al. (2018) for pose transition. Furthermore, Zhang et al. (2019) combined the AutoEncoder and GAN to form an integrated IL-GAN for generate illumination variation. This IL-GAN model has a GAN outline and takes two nested AutoEncoders as the generator. The GANimation Pumarola et al. (2018) model proposed by Pumarola et al. uses a GAN structured network to generate virtual face image conditioning on Action Units (AU) annotations. Two AutoEncoders for regressing attention and color maps make up the generator, while the discriminator evaluates the photo-realism and AU annotations fulfillment.

Moreover, as an emerging research field, face reenactment Garrido et al. (2014) is attracting increasing attention. Face reenactment is to synthesize identity-preserved face images from reference images while transferring the variations from guide images. As a result, it can be also considered an available approach for virtual image generation. The one-shot face reenactment, which refers to performing face reenactment in a single sample situation, was first introduced by Zhang et al. (2019) in 2019. These authors used a shape encoder to extract pose variation from the guide image, along with an appearance encoder to extract identity information from the reference image. The extracted features were then fused by a decoder to generate the virtual samples. Yao et al. (2020) used CNN to regress the 3DMM parameters of the guide image and target image, then estimated the optical flow between them, after which the target image and the optical flow were inputted to an AutoEncoder-based reenacting module. Li et al. (2016) approached face reenactment as an optimization problem, and accordingly generated the target samples by minimizing feature-level attribute loss and identity loss. Kosarevych et al. (2020) used facial landmarks to describe variations in expression and applied a Feature Pyramid Network (FPN) to generate virtual images. Similarly, FaR-GAN Hao et al. (2020) also describes the guided variation using landmarks. The U-Net-structured FaR-GAN fuses the noise and information from reference images and guided landmarks in the up-sampling convolution layers.

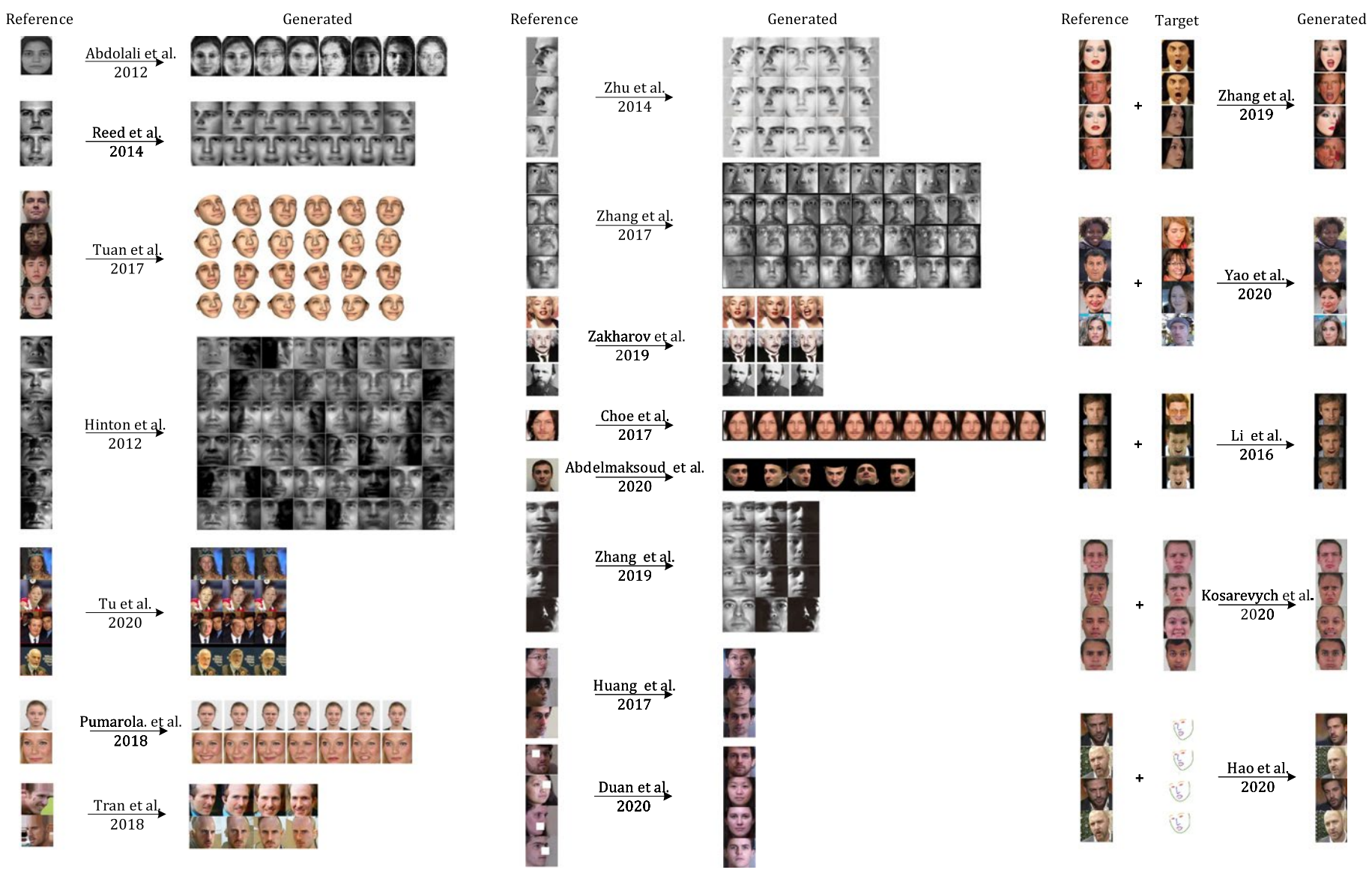

**Fig. 6** Virtual images generated from single sample face images

The generated virtual samples of the above methods are shown in Fig. 6. The first and second columns are corresponding to AutoEncoder-based or GAN-based methods, and the last column is face reenactment methods. Among current virtual sample methods, AutoEncoder and GAN are the most common choices. The AutoEncoder structure expects the model to retain as much information as possible throughout the reconstruction process. While this is conducive to the retention of identity information, it also limits learning and the generation of intra-class variation. Therefore, the generated faces are prone to possess similar features. For its part, GAN can provide stronger intra-class variation generation. However the two-player min-max training means that the model can find it difficult to converge. Unlike virtual image generation methods based on manipulating latent variables, face reenactment can only transfer the variation from another guide image. It is also worth noting that the guide image can interfere with identity information. However, for face reenactment methods, the posed intra-class variations come from real images, so it can reduce the risk of transferring unreal variations to some extent.

### 2.1.2 Virtual feature generation

Virtual image generation methods typically generate images from a latent variable in feature space. The generated image will be further passed to the feature extractor and mapped back to the feature space again. Compared to these methods, generating virtual features can prevent unnecessary information loss caused by the decoding–encoding generation process. Therefore, in addition to the above-mentioned virtual image generation method, a number of virtual feature generation methods have also been proposed. For example, Yin et al. (2019) assumed that the intra-class variations of the feature vectors obeys Gaussian distributions, and accordingly generated multiple virtual features for a single sample by sampling in the corresponding distribution. Min et al. (2019) learned the intra-class variation by clustered features from other persons. This feature expansion method encounters a similar problem with Zhang and Peng (2017) that the similarity between the clustered feature sets might lead to limited intra-class variation generation ability. Moreover, features extracted from different persons cannot characterize the intra-class variation of a single person. Ding et al. (2018) proposed a Conditional Generative Adversarial Network (CGAN)-based virtual feature generation method. Rather than simply assuming that the intra-class variation obeys a certain distribution, these authors used adversarial training to learn the real distribution of the intra-class variation.

However, the above methods regarded the extracted single sample's feature as the center of the corresponding class. They do not consider the variation carried by the single sample. Therefore, in addition to the above feature expansion methods, feature rectification also attracted much attention. Ding et al. (2019) proposed a dictionary-based method, which selects the intra-class variations that are most relevant to the single sample and provides them to the generator. As a more effective approach, feature rectification can directly rectify the variation. Zhou et al. (2020) proposed a Feature Rectification Generative Adversarial Network (FR-GAN) method to reduce the intra-class variation, generating rectified virtual features to train the classifier. Very recently, Pang et al. proposed Variation Disentangling scheme using Variation Disentangling Generative Adversarial Network (VD-GAN) Pang et al. (2021a) and Disentangled Prototype plus Variation model (DisP+V) Pang et al. (2021b) method to generate central images and features for the single sample.

## 2.2 Generic learning methods

Generic learning methods use an additional training set to improve the model performance, and were already a research hotspot before the rise of deep learning. Compared with the methods only using original single sample set, generic learning methods use both the original single sample set and a multi-sample generic set (also known as novel set and base set) to train the model.

In 2006, Wang et al. (2006) proposed the framework of generic learning, which was subsequently refined by Su et al. and Kan et al. through the development of Adaptive Generic Learning (AGL) Su et al. (2010) and Adaptive Discriminant Analysis (ADA) Kan et al. (2013), respectively. Based on Equidistant Prototypes Embedding, Deng et al. (2014) proposed Linear Regression Analysis (LRA) and leveraged generic learning to improve the generalization performance. Sparse representation is also an important branch of generic learning. To solve the performance degradation problem of the Sparse Representation Classifier (SRC) in the single sample situation, Deng et al. (2012) proposed an extended SRC (ESRC) method. Liu et al. (2015, 2016) also proposed local structure-based SRC (LS-SRC) and local structure based multi-phase CRC (LS-MPCRC) to solve this problem. Alongside the rise of deep learning in recent years, researchers have begun to use deep learning to extend traditional generic learning methods. As Fig. 7 shows, some researchers have attempted to combine deep features in order to improve the recognition performance of traditional generic learning methods, while others have focused on improving the deep model from the loss function or network structure aspect.

### 2.2.1 Combining traditional methods and deep features

In an attempt to take full advantage of the useful properties of traditional methods, many researchers have tried to combine traditional methods, such as the sparse classifier and SVM, with features extracted from deep models. For example, Jadhav et al. (2016) used CNN to extract attribute features that describe certain face attributes, such as gender, hair color, and face shape, then combined these attribute features with Exemplar-SVM Malisiewicz et al. (2011) or SVM with one-shot similarity kernel Wolf et al. (2009). Ouanan et al. (2018) proposed a non-linear dictionary representation of deep features and applied Fisher Discrimination Dictionary Learning (FDDL) Yang et al. (2011) to deep features. Adamo et al. (2012) proposed to use the k-LiMapS algorithm for SRC-based face recognition. Bodini et al. (2018) combined this method with deep features, and then Cuculo et al. (2019) further improved the method in the face augmentation and sparse sub-dictionary learning steps. In a similar vein, the Probabilistic Collaborative Representation-based Classifier (ProCRC) Cai et al. (2016), Semi-supervised Sparse Representation ($S^3$RC) Gao et al. (2017), Synergistic Generic Learning (SGL) Pang et al. (2019), Iterative Dynamic Generic Learning (IDGL) Pang et al. (2020), Superposed SRC (SSRC) Deng et al. (2013), Sparse Variation Dictionary Learning (SVDL) Yang et al. (2013), and Collaborative Probabilistic

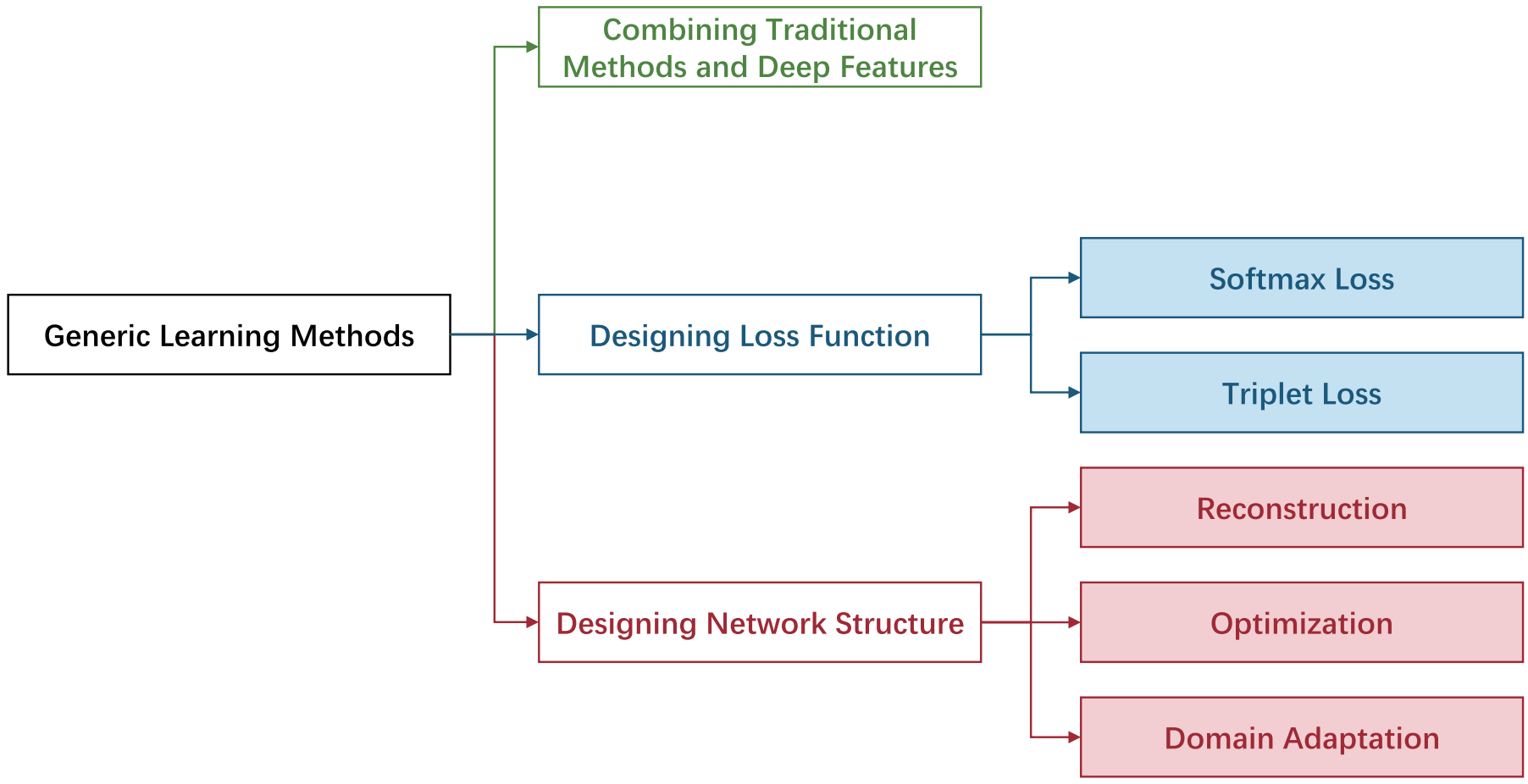

**Fig. 7** The different categories of generic learning methods

Labels (CPL) Ji et al. (2017) have all extended SRC from different perspectives, making attempts to combine the proposed methods with deep features in their experiment sections.

Patch-based methods are also an important branch of the traditional methods. Yang et al. proposed Joint and Collaborative Representation with local Adaptive Convolution Feature (JCR-ACF) Yang et al. (2017). This approach extracts deep feature from local patches, then enforces these features to have similar coefficients in sparse representation. Based on JCR-ACF, these authors further proposed Robust Joint Representation (RJR) Wang et al. (2019) and Class-level Joint Representation with Regional Adaptive Convolution Feature (CJR-RACF) Wen et al. (2018). RJR improves the original method from the sparse representation perspective, while the CJR-RACF makes improvements from the CNN perspective. A shared convolutional layer is employed in CJR-RACF, since the local low-level features of the image share similar patterns. Liu et al. (2019) used the deep features extracted by JCR-ACF and proposed a low-rank regularized generic representation method. Recently, Ding et al. (2020) designed a uniform generic representation approach that integrates global generic representation and local representation. According to the reported experimental results of the above models, the introduction of deep features can improve performance by 30.27% on average. However, existing works of this kind have not improved the deep learning model for single sample face recognition tasks, and have not yet exploited the full potential of deep learning.

### 2.2.2 Improving loss function

As most deep learning models learn parameters by back-propagation according to the loss function, one straightforward approach to make the model adapt to the single sample situation would be improving the loss function. Current loss functions in the single sample face recognition field can be divided into two types: softmax loss and triplet loss. In the following, we will discuss the improvement of the two loss function.

As shown in Fig. 8, the training strategies adopted by softmax loss-based methods can be divided into two types, one is to those that train the model using the generic set and the single sample set in a two-step manner, and the other is to train the model in a single-step manner by fusing the generic set and the single-sample set. Both two types have certain problems, and researchers have proposed various solutions. For the first type, the models often fall into overfitting during fine-tuning due to the small size of the single sample set. In response, Zeng et al. (2017) used the virtual sample method to enlarge the single sample set when fine-tuning. Wu et al. (2017) avoided fine-tuning through the use of a hybrid classifier composed of a softmax classifier and a nearest neighbor classifier. For the second type, the generic set and the single sample set form a fused training set. In practice, researchers have found that this causes the weight norms in the softmax layer of single sample classes to be smaller than those of the multi-sample classes, leading to classification boundary shifted and decision area decreased. Guo et al. refer to these classes as underrepresented classes and proposed a novel loss term, underrepresented classes Promotion (UP) Guo and Zhang (2017), to compensate for the smaller weight norms. Wang et al. (2018) simplified the UP term by proposing to regularize weight norms, further balancing the classification boundaries and reducing the computational cost of training. Cheng et al. (2017) proposed an enforced softmax optimization approach, which sequentially leverages optimal dropout, selective attenuation, L2 normalization, and model-level optimization. Of these, the L2 normalization term rectifies the classification boundary by normalizing the weight norms. These softmax loss-based methods focus primarily on the weight norm decreasing problem caused by the imbalance between the generic set and single sample set. This situation is an extreme case of the long-tailed distribution of the training set; thus, many loss functions that solve the long-tailed distribution problem have the potential to solve single sample problem.

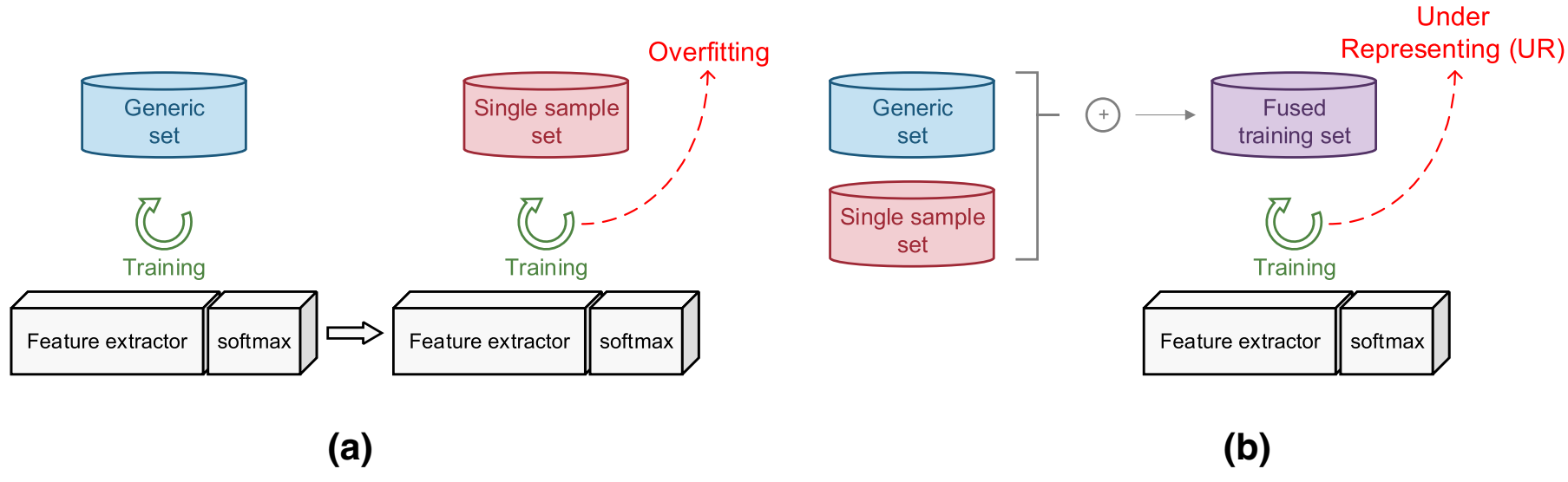

**Fig. 8** **a** 2-step training and **b** single-step training of softmax-loss based methods

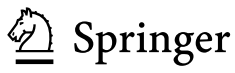

When some face recognition tasks reach tens of thousands of classes, the softmax layer could result in very high computational cost. Triplet loss-based methods can solve this problem to some extent. The diagram of these methods are shown in Fig. 9. Schroff et al. (2015) proposed FaceNet by using triplet-loss to reduce the distance between the anchor-positive sample pairs and increase the distance between anchor-negative pairs. Parchami et al. (2017a) proposed CCM-CNN to further increase the distance between positive-negative pairs, thereby enhancing the discrimination of the extracted face features to a greater extent. Ding and Tao (2017) proposed the Mean Distance Regularized Triplet Loss (MDR-TL) to normalize the inter-class distance in order to improve the model's discriminatory capacity. Based on MDR-TL, Parchami et al. (2017b) further normalized the intra-class distance, obtaining inter-class sparse and intra-class compact face features. In the training of triplet loss-based methods, better triplet selection can both promote convergence and enhance the robustness of the feature extractor. Accordingly, Smirnov et al. (2017) proposed a Doppelganger Mining (DM) method, which uses softmax loss to maintain a list of the identities most similar to each identity in the training set, thereby generating better mini-batches and thus benefiting the training of deep model.

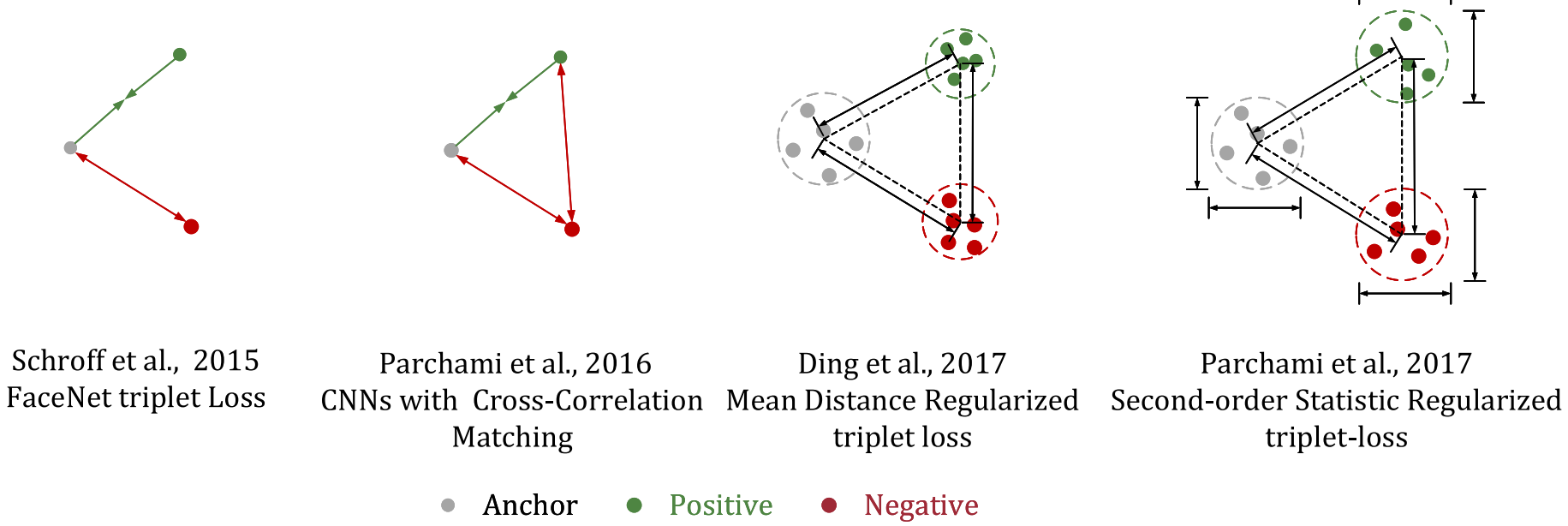

**Fig. 9** Diagrams of triplet loss-based methods proposed by Schroff et al. (2015), Parchami et al. (2016, 2017a, 2017b) and Ding and Tao (2017)

### 2.2.3 Improving network structure

Based on the conventional deep learning model, designing loss function can effectively alleviate the single sample problem. At the same time, a novel network structure can also benefit the model learning the intra-class and inter-class facial variations. Researchers have found that when auto-encoders are applied to reconstructing the face images of the same subject rather than reconstructing the original image, the auto-encoder can obtain face features that are robust to intra-class variation. For example, to obtain Face Identity-Preserving (FIP) features, Zhu et al. (2013) added a normalized face reconstruction task following feature extraction via CNN. However, the model may use only certain parts of the dimensions in the FIP feature for reconstruction, and the other dimensions remain affected by intra-class variation. Gao et al. proposed a Stacked Supervised AutoEncoder (SSAE) Gao et al. (2015). Similar to Zhu et al. (2013), the model reconstructs a normalized face, and additionally enforces the face features extracted from the same subject to be similar, thus minimizing the impact of intra-class variation. Since AutoEncoder with sparsity constraint can usually achieve better performance, the SSAE uses a sparse regularization loss term based on KL divergence. Subsequently, Viktorisa et al. (2016) conducted comparative testing of three forms of sparse regularization terms and confirmed that a regularization term based on KL divergence achieved the best performance. In Vega et al. (2016), researchers trained SSAE using face samples that had been automatically extracted from video data. Compared with the dataset in Gao et al. (2015), the distribution of intra-class variation of face samples in the video data is closer to the distribution in real-world scenarios. Based on the network structure of the FIP method, Wu and Deng (2016) added another reconstruction task that reconstructs the original input face from the normalized face. A network structure aids the model in effectively separating identity information and intra-class variation, thereby improving the feature robustness.

The above methods obtain robust features through the addition of auxiliary reconstruction tasks, reducing the impact of intra-class variation on features. Some researchers have further pointed out that deep learning models can also use optimization methods other than back-propagation to learn parameters. Chan et al. (2015) proposed using PCA to learn convolution kernel parameters, after which Ding et al. (2017) added a local patch-based weighted voting scheme and further improved the model performance. Based on sparse representation, Zhang et al. [82] designed an end-to-end deep cascade model without back-propagation Zhang et al. (2019). In Guo et al. (2017), Guo et al. proposed a one-shot face recognition model based on a sparse auto-encoder. These authors used fuzzy rough set theory to remove redundant parameters. Deep methods of this kind can to some extent alleviate the over-fitting problem caused by the single sample situation. At the same time, novel parameter learning schemes can not only ensure that the model effectively extracts features from the single sample, but also reduce the computational cost associated with training.

In generic learning methods, the deep model learns the representation and discrimination of face samples on the generic set (the source domain) and applies it to the single sample set (the target domain). The difference between the source and target domains could potentially present two key drawbacks. The first is over-adaptation; that is, the deep model overfits the target domain and forgets the knowledge learned from the source domain. The second is under-adaptation, which means that the model fails to make good use of the information in the target domain and is unable to adapt to it. For the first type of problem, You et al. (2017) proposed a transfer learning method based on Restricted Parameter Learning (RPL); the RPL limits the learning of the CNN parameters during training truncates gradients that exceed a certain threshold. For the second type of problem, the SSPP-DAN Hong et al. (2017) applied Domain Adaptation Network (DAN) to the single sample face recognition task, using a domain discriminator to enforce the generation of domain-robust features by the feature extractor.

## 3 Databases and performance comparison

In this section, we review some commonly used databases, along with the corresponding evaluation results of deep learning based single sample face recognition methods. Among existing works, the most commonly used databases are AR Martinez (1998), Extend Yale B Georghiades et al. (2001), LFW Huang et al. (2008), and MS-Celeb-1M Guo et al.

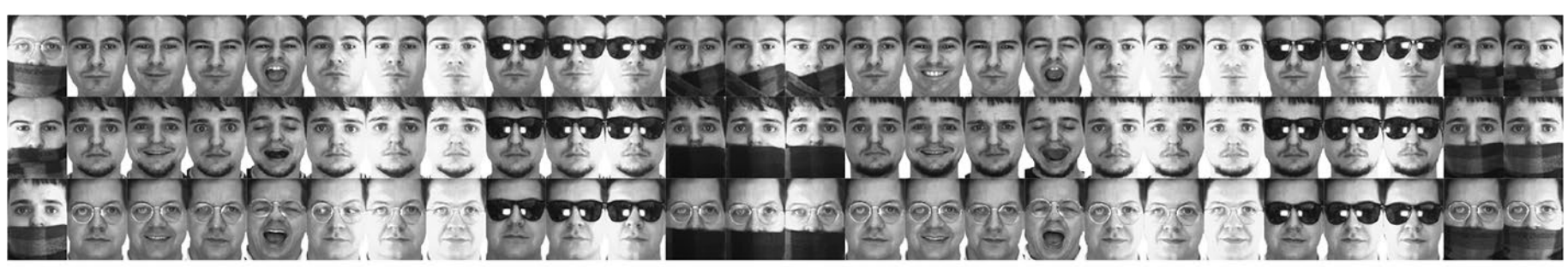

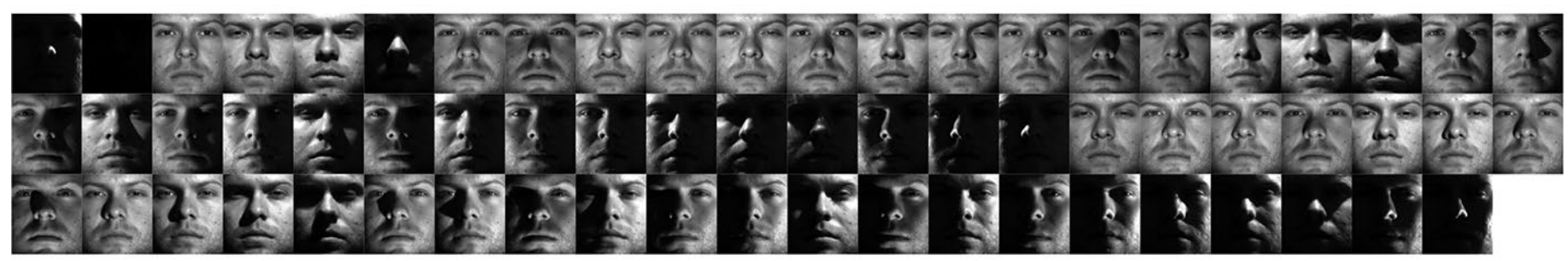

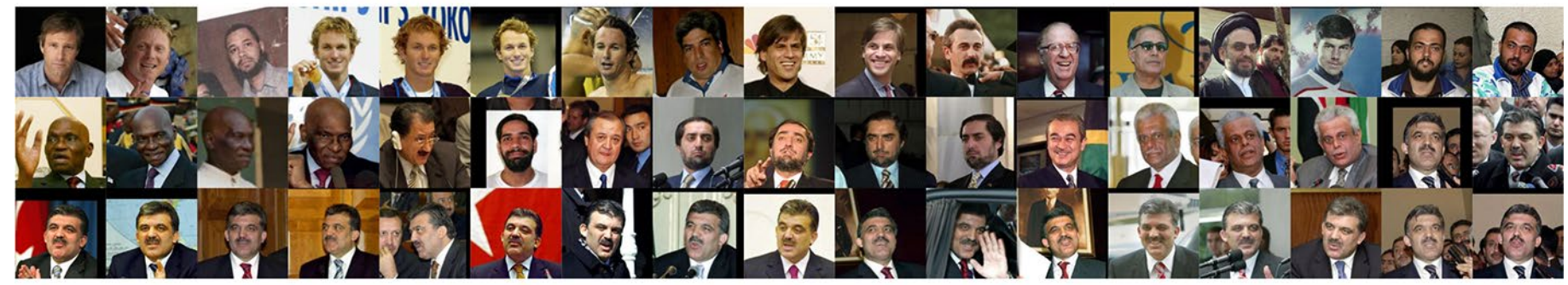

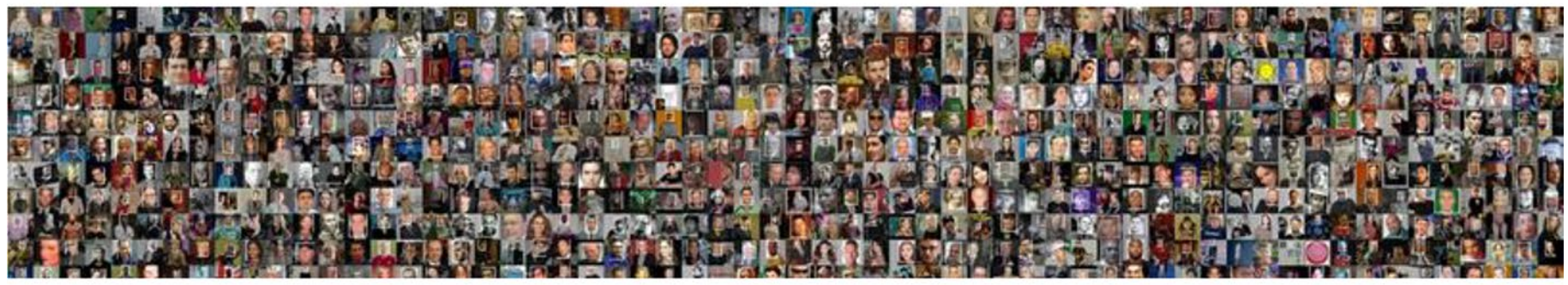

**Fig. 10** Samples of AR, Extend Yale B, Labeled Faces in the Wild (LFW) and MS-Celeb-1M database

(2016),which are shown in Fig. 10. The AR Database Martinez (1998) is a well-known face dataset created in 1998. It contains over 2600 gray-scale images of 100 subjects with 120×165 resolution; images are characterized by different facial expressions, illumination conditions, and occlusions. The Extend Yale B Database Georghiades et al. (2001), created in 2001, is an extension of the Yale Database and Yale B Database. It contains 5760 gray-scale face images of 10 subjects with 168×192 resolution. Labeled Faces in the Wild (LFW) dataset Huang et al. (2008) is a large-scale unconstrained face dataset collected in 2008. It contains 13,233 RGB face images of 5749 subjects with 250×250 resolution. Finally, the MS-Celeb-1M Dataset Martinez (1998) is a large-scale database released by Microsoft in 2016, which contains web-collected RGB images of one million subjects. Each identity in these databases has multiple samples, and these databases are also commonly used in multi-sample face recognition. As described in the literature Philipps et al. (1998), the known face images to an algorithm are referred to as the gallery and the unknown face images to that are referred to as the probe, the collection of galleries and probes are known as gallery set and probe set, respectively . For the SSPP-FR problem, there is only one sample per person in the gallery set. Usually, using the neutral face (frontal face, neutral expression, no illumination and no disguise) as the gallery is the beneficial setting for SSPP-FR, and the number of individuals in the gallery set is the same as that in the probe set Tan et al. (2006).

Same as multi-sample face recognition, most single sample face recognition methods adopted accuracy as the evaluation metric. In Fig. 11, we plotted deep learning based single sample face recognition methods' performance on these four datasets, and further marked the average and the best accuracies. More detailed experimental results are presented in Table 1. Moreover, for some of the models that combine traditional methods and deep features, we also add the results of combining traditional features for ease of comparison. As shown in Table 1, the best accuracy results produced by deep learning based single sample face recognition models on the AR, Extend Yale B, LFW, and MS-Celeb-1M datasets were achieved by Liu et al. (2019) (98.50%), Zhang et al. (2019) (97.80%), Wang et al. (2019) (99.29%), and Cheng et al. (2017) (99.01%), respectively. On the LFW dataset, the uniform generic representation method Ding et al. (2020) only achieves the accuracy of

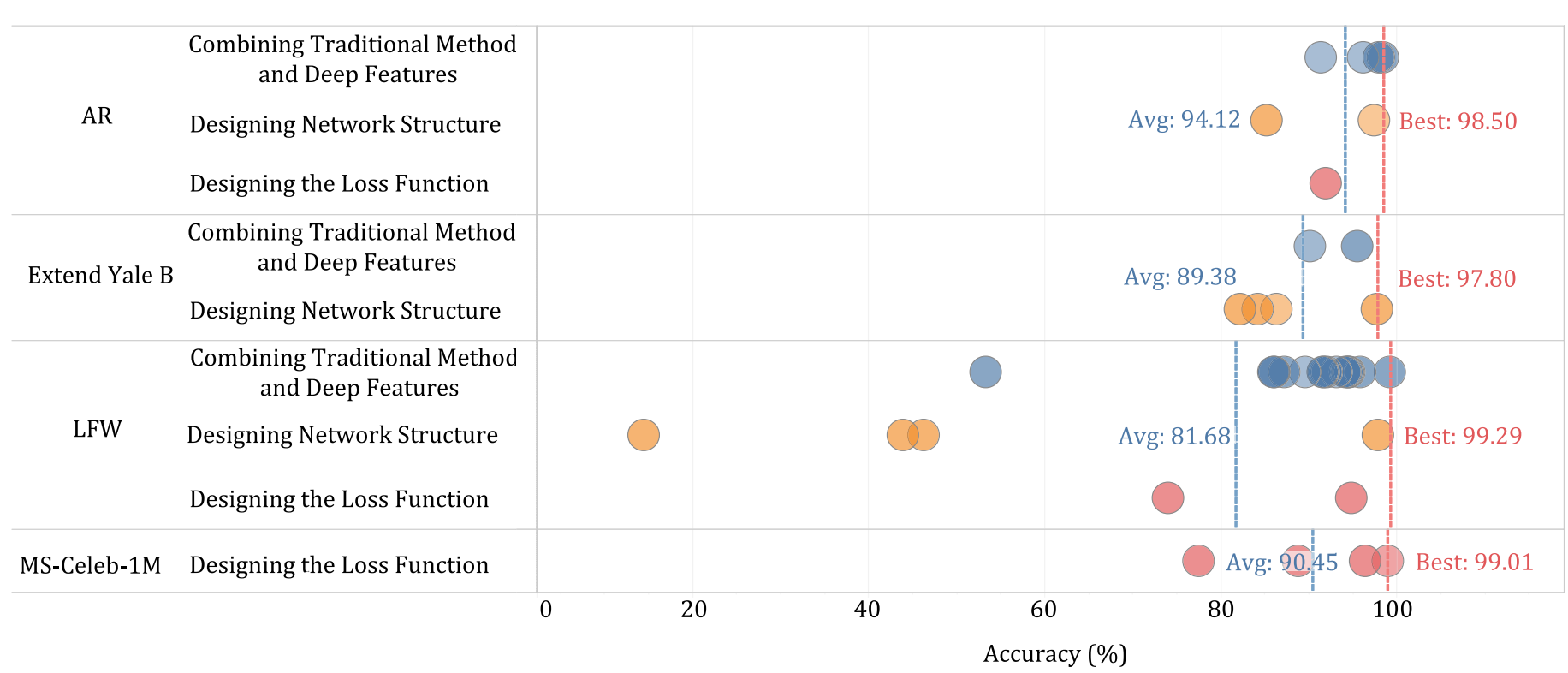

**Fig. 11** Performance evaluation of different single sample face recognition methods on AR, Extend Yale B, Labeled Faces in the Wild (LFW) and MS-Celeb-1M database. Each circle denotes a reported result. Blue: combining traditional method and deep features; Yellow: designing network structure; red: designing loss function

**Table 1** Performance evaluation of different types of one-shot face recognition models covered in this paper. Approaches with the best performance on each dataset are marked in bold

| Database | Type | References | Years | Method | Accuracy (%) | |
|---|---|---|---|---|---|---|
| AR | Combining Traditional Methods and Deep Features | Ouanan et al. (2018) | 2018 | Fisher Discrimination Dictionary Learning (FDDL) | 95 | |
| | | | | Fisher Discrimination Dictionary Learning (FDDL) + Deep Feature | 98 | |
| | | Cuculo et al. (2019) | 2019 | k-LiMapS + Deep Feature | 91.43 | |
| | | Yang et al. (2017) | 2017 | Joint and Collaborative Representation with local Adaptive Convolution Feature (JCR-ACF) | 96.2 | |
| | | Liu et al. (2019) | **2019** | **Low-rank Regularized Generic Representation with Block-Sparse Structure** | **98.5** | |
| | Designing the Loss Function | Zeng et al. (2017) | 2017 | Expanding the fine-tuning training set | 92 | |
| | Designing Network Structure | Gao et al. (2015) | 2015 | Stacked Supervised Auto-Encoders (SSAE) | 85.21 | |
| | | Chan et al. (2015) | 2015 | PCANet | 97.5 | |
| | | Ding et al. (2017) | 2017 | KPCANet with Weighted Voting Scheme (KWV) | 97.5 | |
| Extend Yale B | Combining Traditional Methods and Deep Features | Yang et al. (2017) | 2017 | Joint and Collaborative Representation with local Adaptive Convolution Feature (JCR-ACF) | 90.2 | |
| | | Liu et al. (2019) | 2019 | Low-rank Regularized Generic Representation with Block-Sparse Structure (LRGR-BSS) | 95.5 | |
| | Designing Network Structure | Gao et al. (2015) | 2015 | Stacked Supervised Auto-Encoders (SSAE) | 82.2 | |
| | | Chan et al. (2015) | 2015 | PCANet | 86.4 | |
| | | Ding et al. (2017) | 2017 | KPCANet with Weighted Voting Scheme (KWV) | 84.2 | |
| | | Zhang et al. (2019) | **2019** | **Deep Cascade Model** | **97.8** | |
| LFW | | | | | | Num. subjects |
| | Combining Traditional Methods and Deep Features | Jadhav et al. (2016) | 2016 | SVM with One-shot Similarity kernel | 58.55 | 50 |
| | | | | SVM with One-shot Similarity kernel + Deep Attribute Feature | 86.13 | 50 |

**Table 1** (continued)

| Database | Type | References | Years | Method | Accuracy (%) | |
|---|---|---|---|---|---|---|
| | | Bodini et al. (2018) | 2018 | k-LiMapS + Deep Feature | 95.93 | 50 |
| | | Cuculo et al. (2019) | 2019 | k-LiMapS + Deep Feature | 94.38 | 50 |
| | | Gao et al. (2017) | 2017 | Semi-supervised Sparse Representation ($S^3$RC) | 70.4 | 200 |
| | | | | Semi-supervised Sparse Representation ($S^3$RC) + Deep Feature | 93.2 | 200 |
| | | Pang et al. (2019) | 2019 | Synergistic Generic Learning (SGL) | 24.5 | 80 |
| | | | | Synergistic Generic Learning (SGL)+ Deep Feature | 94.2 | 80 |
| | | Pang et al. (2020) | 2020 | Iterative Dynamic Generic Learning (IDGL) | 77.9 | 200 |
| | | | | Iterative Dynamic Generic Learning (IDGL) + Deep Feature | 94.5 | 200 |
| | | Deng et al. (2012) | 2012 | Extended SRC (ESRC) | 67.4 | 200 |
| | | | | Extended SRC (ESRC) + Deep Feature | 89.6 | 200 |
| | | Deng et al. (2013) | 2013 | Superposed SRC (SSRC) | 69.8 | 200 |
| | | | | Superposed SRC (SSRC) + Deep Feature | 91.7 | 200 |
| | | Yang et al. (2013) | 2013 | Sparse Variation Dictionary Learning (SVDL) | 67.8 | 200 |
| | | | | Sparse Variation Dictionary Learning (SVDL) + Deep Feature | 92.1 | 200 |
| | | Ji et al. (2017) | 2017 | Collaborative Probabilistic Labels (CPL) | 70.9 | 200 |
| | | | | Collaborative Probabilistic Labels (CPL) + Deep Feature | 91.6 | 200 |
| | | Yang et al. (2017) | 2017 | Joint and Collaborative Representation with local Adaptive Convolution Feature (JCR-ACF) | 86 | 158 |
| | | Wang et al. (2019) | **2019** | Robust Joint Representation (RJR) | 50.2 | 158 |
| | | | | **Robust Joint Representation (RJR) + Deep Feature** | **99.29** | 158 |

**Table 1** (continued)

| Database | Type | References | Years | Method | Accuracy (%) | |
|---|---|---|---|---|---|---|
| | | Wen et al. (2018) | 2018 | Class-level Joint Representation with Regional Adaptive Convolution Feature (CJR-RACF) | 94.6 | 158 |
| | | Liu et al. (2019) | 2019 | Low-rank Regularized Generic Representation with Block-Sparse Structure (LRGR-BSS) | 53.3 | 158 |
| | | Ding et al. (2020) | 2020 | Uniform Generic Representation | 46.3 | 158 |
| | | | | Uniform Generic Representation + Deep Feature | 87.3 | 158 |
| | Designing the Loss Function | Zeng et al. (2017) | 2017 | Expanding the fine-tuning training set | 74 | 50 |
| | | Guo and Zhang (2017) | 2017 | Underrepresented-classes Promotion (UP) | 94.89 | 50 |
| | Designing Network Structure | Chan et al. (2015) | 2015 | PCANet | 43.9 | 158 |
| | | Ding et al. (2017) | 2017 | KPCANet with Weighted Voting Scheme (KWV) | 46.3 | 158 |
| | | Guo et al. (2017) | 2017 | Fussy Sparse Autoencoder | 14.52 | 158 |
| | | Hong et al. (2017) | 2017 | Domain Adaption Network | 97.91 | 158 |
| MS-Celeb-1M | Designing the Loss Function | Wu et al. (2017) | 2017 | CNN + Hybrid classifiers | 96.42 | |
| | | Guo and Zhang (2017) | 2017 | Underrepresented-classes Promotion (UP) | 77.48 | |
| | | Wang et al. (2018) | 2018 | Weight norms regularization | 88.87 | |
| | | Cheng et al. (2017) | **2017** | $l_2$ **normalization** | **99.01** | |

For results in LFW dataset, we noted the number of used samples in the last column

46.3%, but it gained 41% improvement from combining deep features, and the SVM with one-shot similarity kernel method Wolf et al. (2009) has 27.58% improvement after using deep learning features. Moreover, the low-rank regularized generic representation with block-sparse structure method Liu et al. (2019) and RJR method Wang et al. (2019) that both combined traditional methods with deep features, achieve the highest accuracy on the AR dataset and LFW dataset respectively. Since AR dataset and LFW dataset both contain complex facial variations, it can be demonstrated that combining traditional methods with deep learning methods can achieve greater advantages in the task of single sample face recognition.

# 4 Analysis and discussion

## 4.1 Identity information preservation of virtual samples

Modern GANs have achieved impressive performances on generating realistic face images Karras et al. (2021), but it doesn't mean that they naturally become perfect choices for virtual sample-based single sample face recognition. Since the face recognition models are directly trained to classify the generated face samples, the model performance would be harmed if the identity information in these generated samples is changed. Therefore, identity information preservation is crucial for virtual sample-based methods. Currently, for GAN-based approaches, identity information preservation can be done in mainly two ways. The first one is CGAN-based, i.e., feeding the identity label to generator and/or discriminator Tran et al. (2018). However, it can not handle the open-set generation scenarios, making it not applicable for single sample face recognition. The second is in a perceptual loss-like fashion (Huang et al. 2017; Duan and Zhang 2020), i.e., matching the features extracted by a pre-trained face model. It is worth noting that the identity preservation ability is upper-bounded by the recognition capacity of the pre-trained face model. In addition to GAN-based methods, for the AutoEncoder-based methods that use the "encode—separate and adjust latent variables—decode" generation process, the identity information can also be affected when the latent variables are adjusted incorrectly.

It also should be noted that, though a lot of empirical works (Mokhayeri and Granger 2020; Tang et al. 2012; Tu et al. 2020; Zhu et al. 2014; Zakharov et al. 2019; Choe et al. 2017; Abdelmaksoud et al. 2020; Feng et al. 2018; Zhang et al. 2019; Pumarola et al. 2018) have proved the effectiveness of virtual sample method, there are few corresponding theoretical guarantees. Many open questions need in-depth investigation. For example, on what condition the virtual samples are helpful? How many virtual samples are enough for learning a good face model? More importantly, how does these virtual samples methods actually work? Virtual sample generation has been be regarded as a certain type of data augmentation, but it is very different from basic image transformations (e.g., random cropping, resizing, etc). For example, GAN-based virtual sample methods can be viewed as transferring GAN's knowledge to the face recognition model. However, the relationship between properties of the GAN's training set (such as scale, diversity) and the final performances of the face recognition model should be further investigated. Finally, we also note that novel metrics that can evaluate the quality and the effectiveness of generated virtual samples will benefit the community, since it is hard and unreliable to compare different virtual sample methods based on subjective judgment only.

### 4.2 Domain adaptation in generic learning

As generic learning methods introduce additional multi-sample generic set to train the face recognition model, there might be a large domain gap between the generic set and single sample set. Such domain gap can be caused by different image collection condition (e.g., pose, light, image resolution) or different face domain (e.g., age, race, expression). In many real world face recognition applications (e.g., surveillance-based, UAV-based, or mobile phone-based face recognition), target faces are usually in different domains compared to pre-training face datasets. In such situation, face feature learned from generic set might be less useful in target face images. For example, it might not be the best choice to directly applying LFW-trained face feature to a kindergarten single sample face recognition. This leads to the necessity of domain adaptation, which is a important branch of transfer learning. Domain adaptation is a learning technique to address the problem of lacking massive amounts of high-quality, large-scale labeled training data in certain target domain. With the continuous development of deep learning, investigating deep domain adaptation has emerged as a new research direction. In recent years, there are lots of advances on deep visual domain adaptation Wang and Deng (2018), but few of them are for single sample face recognition. As in Sect. 2.2.3, there are only two existing domain adaptation methods (You et al. 2017; Hong et al. 2017) for single sample face recognition to our best knowledge. Therefore, many other domain adaptation methods Wang and Deng (2018) can be further validated for single sample face recognition.

The domain problem also appears in a specific type of face recognition: heterogeneous face recognition, which includes visible-near infrared matching (He et al. 2021; Wu et al. 2019), sketch-to-photo matching (Fan et al. 2021; Yu et al. 2019; Liu et al. 2020; Galea and Farrugia 2017a,2017b; Klum et al. 2014), etc. In some heterogeneous face recognition tasks, such as in criminal investigation scenario, there is only a single query face image available. Though, in such situation, these heterogeneous face recognition can be considered as a special type of single sample face recognition, most works in this field aim at solving the cross-domain matching problem, such as extracting domain-invariant face feature (Klare et al. 2011; Zhang et al. 2011; Gong et al. 2017), learning common feature space (Klare and Jain 2013; Sharma and Jacobs 2011; Li et al. 2009). It should be noted that synthesize-based methods (Gao et al. 2012; Wang et al. 2018; Cao et al. 2019) have shown great success in heterogeneous face recognition recently. These methods convert the heterogeneous problem into a homogeneous one by doing cross-modal translation. From a domain adaptation perspective, these methods can be categorized into generative domain adaptation Wang and Deng (2018). However, generative domain adaptation methods typically attempt to solve a more complex problem than the recognition task itself (Han et al. 2012; Zhang et al. 2011). Their computational efficiency is usually less than the extracting domain-invariant face feature method and learning common feature space method. Moreover, the performance of the former also depends on the quality of the synthesized samples Galea and Farrugia (2017a).

### 4.3 Learning face representation by large-scale unsupervised pre-training

Deep learning-based face recognition models have far more trainable parameters than traditional face models (Tan et al. 2006; Kumar and Garg 2019). This increases the learning ability of deep models and enables them to handle diverse face variations (e.g., occlusion, angle, illumination). However, one important prerequisite of effective training of these deep models is the existence of a large amount of training data. For deep learning models, more training data usually leads to better performances. But in the single sample face recognition scenario, the single sample set is extremely insufficient for training deep models. To solve this issue, most existing methods in this field introduce additional training datasets. However, it should be noticed that both virtual sample methods and generic learning methods need the annotation of identity labels in the additional training datasets.

Since the fundamental purpose of introducing these additional datasets is to learn better face representations, a question can be raised here: is identity annotation indispensable for learning good face representations? Given the success of recent unsupervised/self-supervised representation learning (Chen et al. 2020; He et al. 2020; Grill et al. 2020; Ericsson et al. 2021), large-scale unsupervised pre-training could be a very promising way to obtain robust face representation without any human annotations of identity labels. For example, with advanced face detection techniques Minaee et al. (2021), we can easily collect a huge amount of unlabelled faces (e.g., from surveillance or web data). As the data collection is fully automated and there is no longer any limitation from human annotation labor, such dataset can have a significantly large scale than existing annotated datasets. The collected dataset accordingly consists of more diverse face variations, which is beneficial for learning robust face features. Therefore, unsupervised pre-training (e.g., contrastive learning) on such collected large-scale dataset has great potential for obtaining powerful face representations. Therefore, large-scale unsupervised pre-training for single sample face recognition would also be a promising future research direction.

### 4.4 Semantic gap

Although the virtual samples or generic sets can characterize the intra-class variation and improve the recognition performance to some extent, they cannot fundamentally solve the semantic gap problem in the single sample situation. When learning face recognition, current deep models typically rely on large-scale labeled training sets, but the human brain seems to operate differently: we are used to using semantic transfer modes. A hierarchical relationship exists between different semantics. Low-level semantics are combined to form high-level semantics, which can in turn be combined to form higher-level semantics. However, current deep learning based single sample face recognition models lack explicit semantic transfer in the learning process.

In some harsh environments, many of the best visual systems are still unable to compete with human visual systems. Therefore, researchers have paid increasing attention to learning from physiology and cognitive science, and have proposed a variety of models based on the theory of visual perception. The current deep learning approaches are also developed from simulating the connection mechanism of human brain neurons. However, most current deep learning based single sample face recognition methods fail to consider semantic information such as gender, age, and ethnicity, along with the semantics implied in the face image, which has led to a lack of image recognition capabilities with semantic understanding. Accordingly, to solve the challenge of single sample face recognition, one worthy avenue of in-depth study would be to learn from the semantic understanding mechanism of the human visual perception system and develop deep learning methods from a semantic perspective. This would also furnish researchers with a more in-depth understanding of human visual perception.

## 5 Conclusion

This paper has presented a comprehensive survey of deep learning based single sample face recognition. Existing methods of this kind can be classified into virtual sample methods and generic learning methods. Virtual sample methods generate virtual face images or virtual face features to expand the training set. They are straightforward and intuitive for building the training pipeline. Identity preservation is important for this type of method, as any variation of identity information will impede the training of the follow-up face recognition model. Generic learning methods use an additional dataset to improve the deep model. Researchers have taken various approaches to improve the generic learning models, which contain combining deep features with traditional methods, as well as making numerous improvements to the loss function and network structure. While significant efforts have been expended in this field in recent years, we point out that the domain adaptation problem in particular merits further research. We also point out that large-scale unsupervised pre-training for single sample face recognition would also be a promising future research direction. From a higher-level perspective, several problems encountered by different single sample face recognition methods can be attributed to the semantic gap. Therefore, it would be worthwhile to conduct in-depth research into deep learning based single sample face recognition methods from the perspective of the semantic gap.

**Acknowledgements** This work was partially funded by Natural Science Foundation of Jiangsu Province under Grant No. BK20191298, Research Fund from Science and Technology on Underwater Vehicle Technology Laboratory (2021JCJQ-SYSJJ-LB06905), Water Science and Technology Project of Jiangsu Province under Grant Nos. 2021072, 2021063.

## Declarations

**Conflict of interest** The authors declare that they have no conflict of interest.